\documentclass[runningheads]{llncs}
\usepackage{graphicx}
\usepackage{mathtools}
\usepackage{tikz}
\usepackage{comment}
\usepackage{amsmath,amssymb} 
\usepackage{color}

\usepackage[capitalize]{cleveref}
\crefname{section}{Sec.}{Secs.}
\Crefname{section}{Section}{Sections}
\Crefname{table}{Table}{Tables}
\crefname{table}{Tab.}{Tabs.}

\usepackage{tikz}
\usepackage{amsmath}

\newcommand{\RB}{\mathbb{R}}
\newcommand{\x}{\textbf{x}}

\usepackage{amssymb}

\usepackage{pifont}
\usepackage{multicol}
\usepackage{multirow}
\usepackage[accsupp]{axessibility}  

\usepackage[usestackEOL]{stackengine}
\usepackage{subfig}

\begin{document}
\pagestyle{headings}
\mainmatter
\def\ECCVSubNumber{7780}  

\title{A Learnable Radial Basis Positional Embedding for Coordinate-MLPs} 



\titlerunning{A Learnable Positional Embedding for Coordinate-MLPs}
%
\author{Sameera Ramasinghe   \and
Simon Lucey }
\authorrunning{Sameera Ramasinghe and Simon Lucey}
%
\institute{Australian Institute for Machine Learning\\
University of Adelaide\\
\email{sameera.ramasinghe@adelaide.edu.au}}

\maketitle

\begin{abstract}
   We propose a novel method to enhance the performance of coordinate-MLPs by learning instance-specific positional embeddings. End-to-end optimization of positional embedding parameters along with network weights leads to poor generalization performance. Instead, we develop a generic framework to learn the positional embedding based on the classic graph-Laplacian regularization, which can implicitly balance the trade-off between memorization and generalization. This framework is then used to propose a novel positional embedding scheme, where the hyperparameters are learned per coordinate (\textit{i.e} instance)  to deliver optimal performance. We show that the proposed embedding achieves better performance with higher stability compared to the well-established random Fourier features (RFF). Further, we demonstrate that the proposed embedding scheme yields stable gradients, enabling seamless integration into deep architectures as intermediate layers. 
\dots
\keywords{Coordinate-networks, positional emebddings, implicit neural representations}
\end{abstract}

\section{Introduction}
 \vspace{-0.6em}
Encoding continuous signals as weights of neural networks is now becoming a ubiquitous strategy in computer vision due to numerous appealing properties. First,  signals in computer vision are typically treated in a discrete manner, \textit{e.g.}, images, voxels, meshes, and point clouds. In contrast, neural representations allow such signals to be treated as continuous over bounded domains, a major virtue being smooth interpolations to unseen coordinates \cite{niemeyer2020differentiable,saito2019pifu,sitzmann2019scene,mildenhall2020nerf,zhong2019reconstructing}. Second, neural representations are differentiable and generally more compact compared to canonical representations \cite{sitzmann2020implicit,dupont2021coin}. For instance, a 3D mesh can be represented using a multi-layer perceptron (MLP) with only a few layers \cite{takikawa2021neural,yu2021plenoctrees}. Consequently, this recent influx of neural signal representations has supplemented a variety of computer vision tasks including images representation \cite{nguyen2015deep,stanley2007compositional}, texture generation \cite{henzler2020learning,oechsle2019texture,henzler2020learning,xiang2021neutex}, shape representation \cite{chen2019learning,deng2020nasa,tiwari2021neural,genova2020local,basher2021lightsal,mu2021sdf,park2019deepsdf}, and novel view synthesis \cite{mildenhall2020nerf,niemeyer2020differentiable,saito2019pifu,sitzmann2019scene,yu2021pixelnerf,pumarola2021d,pumarola2021d,rebain2021derf,martin2021nerf,wang2021nerf,park2021nerfies}. 
 
\begin{figure}[!ht]
    \centering
        \subfloat{\includegraphics[width=0.3\columnwidth]{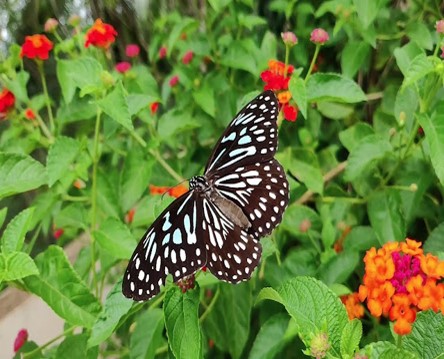}}
        \subfloat{\includegraphics[width=0.3\columnwidth]{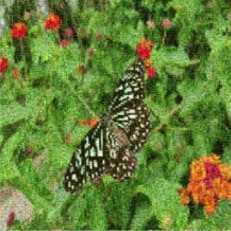}}
        \subfloat{\includegraphics[width=0.3\columnwidth]{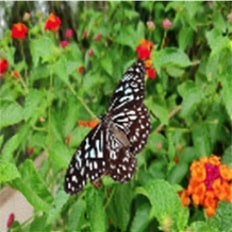}}
\qquad
        \subfloat[Ground truth]{\includegraphics[width=0.3\columnwidth]{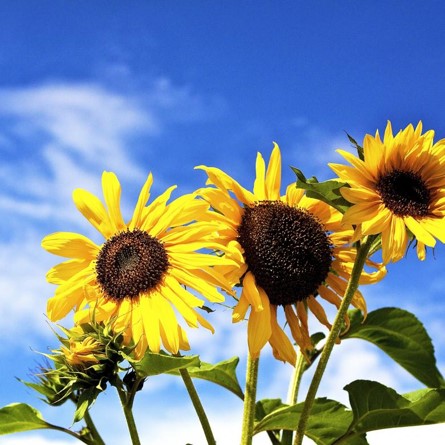}}
        \subfloat[End-to-end training]{\includegraphics[width=0.3\columnwidth]{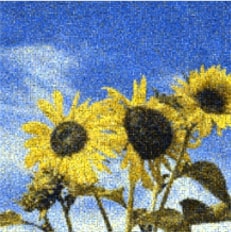}}
        \subfloat[Proposed training]{\includegraphics[width=0.3\columnwidth]{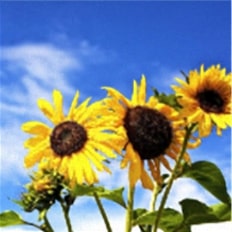}}
  
            \caption{Optimizing the positional embedding layer by minimizing the training error by updating the embedding hyperparameters and the network weights (end-to-end optimization) leads to overfitting, yielding poor generalization. In contrast, the proposed optimization procedure achieves a better trade-off between memorization and generalization. We use ($25\%$)  regularly sampled training points for this example.}
    \label{fig:front_fig}
\end{figure}




The elementary units of neural signal representations are coordinate-MLPs. In particular, a coordinate-MLP consumes a discrete set of coordinates as inputs $\textbf{x}$ and corresponding targets $\textbf{y}$, and strives to learn a continuous approximation $f:\RB^M \to \RB^N$. However, it has been both empirically \cite{mildenhall2020nerf,niemeyer2020differentiable} and theoretically \cite{tancik2020fourier} observed that low-dimensional coordinate inputs (typically $2$ or $3$ dimensional) are sub-optimal for such tasks, since conventional MLPs, when trained with such inputs, fail to learn functions with high-frequency components. A popular approach to tackle this obstacle involves projecting the coordinates into a higher dimensional space via a \emph{positional embedding} prior to the MLP. Positional embeddings -- like MLPs -- have parameters that can be tuned to match the signal. Unlike MLPs, however, they cannot be learned end-to-end; when they are, it can lead to catastrophic generalization/interpolation performance (see Figure~\ref{fig:front_fig}). To circumvent this issue, most methods instead tie the positional embedding parameters across coordinates (dramatically reducing the search space, often to a single parameter), and then select the reduced parameter set through a cross-validation procedure. This inevitably leads to sub-optimal results, as the optimal selection varies from coordinate-to-coordinate and signal-to-signal~\cite{tancik2020fourier,zheng2021rethinking}. Philosophically, this approach is questionable: Modern deep learning has heavily advocated for the learning of parameters (even those in positional embedding) from data. Our paper presents a way forward on how to effectively learn instance (\textit{i.e.} coordinate) specific positional embeddings for coordinate-MLPs.  

 Why does the optimization of positional embedding parameters in Figure~\ref{fig:front_fig} -- using gradient descent to minimize the training error -- catastrophically fail? We postulate that the root cause of this behavior extends to the inherent trade-off between the memorization (of seen coordinates) and generalization (to unseen coordinates) of the embedding. Unlike modern MLPs, the generalization performance of positional embedding must be explicitly -- rather than architecturally -- controlled. In this sense, the values of positional encoding should be treated to control the learning process (\textit{i.e.} hyperparameters) rather than learned through conventional end-to-end training. Our paper proposes that this can be done by optimizing a proxy measure of generalization: namely smoothness. We show that the desired level of smoothness varies locally and equivalently, the distortion of the manifold formed by the positional embedding layer should be smooth with respect to the first-order gradients of the encoded signal. Then, for practical implementation, we extend our analysis to its discrete counter-part using similarity graphs and link the above objective to the graph-Laplacian regularization.

\noindent \textbf{Contributions:} We demonstrate the effectiveness of our approach across a number of signal reconstruction tasks in comparison to leading hard coded methods such as Random Fourier Frequencies (RFF) made popular in~\cite{mildenhall2020nerf}. A unique aspect of our method is that we do NOT use Fourier positional embeddings in our approach. Such embeddings are actually problematic when attempting to enforce local smoothness. Inspired by~\cite{zheng2021rethinking}, we instead opt for embeddings that have spatial locality such as radial basis functions. To our knowledge, this is the first time that: (i) non-Fourier, and (ii) learned hyper-parameters are used to obtain state-of-the-art performance in coordinate-MLPs. We also show that our embedder allows more stable back-propagations through the positional embedding layer, enabling hassle-free integration of the positional embedding layer as an intermediate module in deep architectures. It is also noteworthy that our work sheds light on a rather debatable topic -- deep networks can benefit from non-end-to-end learning, where each layer can be optimized independently with different optimization goals. However, a broader exposition of this topic is out of the scope of this paper.

 \vspace{-0.6em}
\section{Related works}
 \vspace{-0.6em}
Positional embeddings are an integral component of coordinate-MLPs that enable them to encode high-frequency functions with fine details. To this date, random Fourier features (RFF) have been the prevalent method of embedding positions.  One of the earliest works that employed RFF for embedding positions extends back to the preliminary work by Rahimi \textit{et al.} \cite{rahimi2007random}, who found that an arbitrary stationary kernel can be approximated using sinusoidal input mappings. Their work was primarily inspired by Bochner’s theorem.  Such embeddings recently gained momentum after the seminal work by Mildenhall \textit{et al.} \cite{mildenhall2020nerf}, where they used random Fourier features to successfully synthesize novel views of a 3D scene. Since then, a large and growing body of literature has investigated the effects of modulating low dimensional inputs with Fourier features \cite{xu2021h,jain2021putting,deng2021depth,yen2020inerf,trevithick2020grf,bi2020neural,guo2020object}. These works provide significant heuristic evidence that such pre-processing allows MLPs to regress functions with high-frequency content. A recent follow-up theoretical examination by Tancik \textit{et al.} \cite{tancik2020fourier} further confirmed that Fourier embeddings allow tuning the bandwidth of the neural tangent kernel (NTK) of MLPs, enabling them to learn high frequency content. In contrast, Zheng \textit{et al.} \cite{zheng2021rethinking} proposed a novel embedding scheme that involves non-Fourier functions, where they construct the embeddings via samples of shifted basis functions. Despite the efficacy of their method in encoding 1D signals, extending the Gaussian embedder to encode 2D or 3D signals necessitates sampling 2D or 3D Gaussian signals on a dense grid. Thus, the dimension of the embedded coordinates grows exponentially with the dimension of the encoded signal, which is a significant drawback compared to RFF.

\noindent \textbf{Learned embeddings:} Positional embeddings have enjoyed data-driven optimizations in language and sequence modeling. For instance, Shaw \textit{et al.} \cite{shaw2018self} proposed a relative position representation to reduce the number of parameters. Similarly, Lan \textit{et al.} \cite{light} and Dehghani \textit{et al.} \cite{dehghani2018universal} proposed to inject relative position information to every layer of a transformer for improved performance. In contrast, Wang \textit{et al.} \cite{wang2019encoding}  proposed to encode positions
by complex numbers of different phases and wave-lengths. A major drawback of their method was that it increases the size of the embedding matrix by a factor of three. Although the aforementioned methods use priors from data to enhance the embeddings, they are not automatically learnable from data. Liu \textit{et al.} \cite{liu2020learning}, on the other hand, introduced a learnable positional encoding scheme, where they model the evolution of encoded results along position index by a dynamical system. In order to model this dynamical system, they utilize neural ODEs \cite{chen2018neural}. However, all of the above methods focus on embedding word positions when modeling sequential data. In contrast, our work focuses on embedding coordinates to encode continuous signals using MLPs.

The rest of the paper is organized as follows: In Section \ref{sec:methodology}, we formulate our objective by preserving the local smoothness of the embedding from a continuous-domain perspective. In Section~\ref{sec:discrete}, we extend the analysis to its discrete counter-part for practical implementation and link the above objective to the classic graph Laplacian regularization. Section \ref{sec:supergaussian} proposes a novel non-Fourier positional embedding scheme that utilizes the proposed optimization strategy. Finally, in Section \ref{sec:experiments}, we experimentally validate the proposed method across a variety of tasks.

 \vspace{-0.6em}
\section{Smoothness of the positional embedding}
\label{sec:methodology}
 \vspace{-0.6em}
MLPs naturally preserve smoothness when trained end-to-end  to fit a signal \cite{tancik2020fourier,rahaman2019spectral}. That is, they do not tend to critically overfit when encoding functions with high frequencies. On the other hand, positional embeddings demonstrate quite the opposite characteristics by rapidly overfitting to the training inputs when optimized similarly to the MLP weights. Therefore, a crucial aspect that should be considered in any optimization scheme, that attempts to learn positional embeddings, is preserving a form of smoothness with respect to the targets. Consider a simple signal $f(x):\RB \to \RB$. We opt to preserve, 

\begin{equation}
    \frac{\|\Phi(x + \Delta x) - \Phi(x)\|_{2}}{|f(x + \Delta x) - f(x)|} = K
\end{equation}

where $\Phi(\cdot):\RB \to \RB^D$ is the positional embedding. With $\underset{\Delta x \to 0}{\lim}$ we get,

\begin{equation}
    \frac{\| d\Phi(x) \|_{2}}{|dx|} \propto \frac{| df(x) |}{|dx|}
\end{equation}

Assume that $\Phi(\cdot)$ has nonvanishing Jacobian almost everywhere, so that its image in $\RB^D$ is a manifold with Riemannian metric inherited from $\RB^D$. Then, $\|d\Phi(x)\|_2$ can be written in terms of the Riemannian metric tensor $\mathcal{M}$ which gives,

\begin{equation}
    \frac{| \sqrt{\mathrm{det}(\mathcal{M}(x))}dx |}{|dx|} \propto \frac{| df(x) |}{|dx|},
\end{equation}

where $\sqrt{\mathrm{det}(\mathcal{M}(x))}$ is the volume element of $\mathcal{M}$ at $x$. It follows that,

\begin{equation}
    \sqrt{\mathrm{det}(\mathcal{M}(x))}  \propto |\frac{ df(x)}{dx}|.
\end{equation}

\noindent \textbf{Multi-dimensional signals:} When encoding vector valued functions $f(\x): \RB^N \to \RB^{M}$, we approximate the above objective as,

\begin{equation}
    \sqrt{\mathrm{det}(\mathcal{M}(\x))}  \propto \|\textbf{J}_{f}(\x)\|_{F}, \label{equ:prac_condition}
\end{equation}

where $\textbf{J}_{f}(\x) = \frac{\partial f(\x)}{\partial \x}$ is the Jacobian of $f$ at $\x$ and $\| \cdot\|_F$ is the Frobenius norm. This portrays that the volume element of the manifold induced by the positional embedding layer should be proportional to the norm of the first-order gradients (Jacobian) of targets. Intuitively, the goal of Eq.~\ref{equ:prac_condition} can be perceived as follows: $\mathrm{det}(\mathcal{M}(\x))$ is a measure of distortion of a manifold with respect to its coordinate-chart at $\x$. Hence, at intervals where $f(\x_i)$ has high gradients with respect to the coordinates, the manifold will consist of higher distortions. Therefore, the distance between two points on the manifold will be stretched, and the gradients of the outputs with respect to the manifold will be encouraged to remain approximately constant. 

However, a critical obstacle that hampers direct utilization of the above analysis is that, in practice, we are only armed with discrete training samples within a bounded space in $\RB^{N}$. Therefore, it is vital to investigate a scheme that can approximate \ref{equ:prac_condition} using a discrete representation. In Section~\ref{sec:discrete}, we seek to accomplish this task using similarity graphs.

 \vspace{-0.6em}

\section{Discrete approximation of the manifold}
\label{sec:discrete}
 \vspace{-0.6em}
 A graph can be considered as a discrete approximation of a manifold \cite{zhou2008high}. 
 In the \emph{graph signal processing} (GSP) literature, \emph{exemplar functions} are used to lift a low dimensional signal to a higher-dimensional space, which is then used to construct an undirected similarity graph. We observe that positional embedders perform a similar task, \textit{i.e.} project low-dimensional input coordinates to a higher-dimensional space. Following these insights, we construct an undirected similarity graph $\mathcal{G}$ using the positional embeddings with the intention of approximating \ref{equ:prac_condition} with a discrete domain formulation.


Consider a set of coordinates $\{ \textbf{x}_i \}_{i=1}^L$ and corresponding sampled outputs $\{f(\textbf{x}_i)\}_{i=1}^L$, $\textbf{x} \in \RB^N, f(\textbf{x}) \in \RB^M$. Then, with an  embedding function $\phi:\RB^N \to \RB^D$  defined on a bounded interval in $\RB^N$, we obtain a set of $D-$dimensional vectors $\{ \Phi(\textbf{x}_i)\}_{i=1}^L$, where $\Phi(\textbf{x}_i) = [ \phi_1(\textbf{x}_i), \phi_2(\textbf{x}_i), \dots   \phi_D(\textbf{x}_i)]^T$. 
Treating this set of vectors $\{ \Phi(\x_i)\}_{i=1}^L \in \mathcal{V}$ as vertices, we build an undirected graph $\mathcal{G}$ with edges $\mathcal{E}$, where the edge weight $w_{ij}$ between two vertices $\Phi(\x_i),\Phi(\x_j) \in \mathcal{V}$ is defined as
\vspace{-0.5em}
\begin{equation}
    w_{ij} = (\rho_i \rho_j)^{-\lambda} \theta(d_{i,j}),
\end{equation}

Such that
\vspace{-1em}
\begin{align}
\label{equ:graph_smooth}
    \theta(d_{i,j}) = \exp{(- \frac{d^2_{i,j}}{2 \epsilon^2})},
\end{align}

where $d_{i,j} = \|\Phi(\x_i) - \Phi(\x_j)\|_2$ and $\rho_i = \sum_{j=1}^{L} \theta(d_{i,j})$. Further, $\epsilon$ and $\lambda$ are hyper-parameters (we fix these parameters across all our experiments,  hence, hyperparameter sweeps are not required). The term $(\rho_i \rho_j)^{-\lambda}$ normalizes the edge weights with respect to the degree of the corresponding vertex. Likewise, Eq.~\ref{equ:graph_smooth} smooths out the edge weights while suppressing weights between vertices that are too distant.  Finally, the unnormalized graph Laplacian $\textbf{L}$ of $\mathcal{G}$ can be computed as
\vspace{-0.4em}
\begin{equation}
    \mathbf{L} = \mathbf{A} - \mathbf{D},
\end{equation}

where $\mathbf{A}$ and $\mathbf{D}$ are the adjacency matrix and the degree matrix of $\mathcal{G}$, respectively. In Section~\ref{sec:glr}, we discuss the impact of Laplacian regularization of $\mathcal{G}$ on our original objective~\ref{equ:prac_condition}. 

\noindent \textbf{Graph Laplacian:}
\label{sec:glr}
The regularization objective is defined as
\begin{align}
\label{equ:glr}
    \tau (\textbf{u})  = \mathbf{u}^T \mathbf{L} \mathbf{u},
\end{align}

where $\textbf{u} = [\varphi(\textbf{x}_1), \varphi(\textbf{x}_2),\ \dots, \varphi(\textbf{x}_L)]^T$ and $\varphi: \RB^N \to \RB$ is an arbitrary smooth function defined on the coordinate space. Minimizing \ref{equ:glr} encourages $w_{i,j}$ to be larger if $\varphi(\textbf{x}_i)$ and $\varphi(\textbf{x}_j)$ are similar (Appendix). As the distance between inputs approaches zero, Eq.~\ref{equ:glr} converges to its continuous counter-part, \textit{i.e.} \emph{anistropic Dirichlet energy} \cite{pang2017graph}. Further, minimizing the anistropic Dirichlet energy is equivalent to enforcing the quantity $\mathrm{det}(\textbf{G}(\x))$ to be smooth with respect to $\textbf{u}$ \cite{alliez2007voronoi} where

\[
\scriptsize
\label{equ:metric}
  \mathbf{G}(\textbf{x}) = \sum_{i=1}^D  \begin{bmatrix}
      (\frac{\partial \phi_i(\x)}{\partial x_1})^2 & \frac{\partial \phi_i(\x)}{\partial x_1} \frac{\partial \phi_i(\x)}{\partial x_2} & \dots &       \frac{\partial \phi_i(\x)}{\partial x_1} \frac{\partial \phi_i(\x)}{\partial x_N} \\
      \frac{\partial \phi_i(\x)}{\partial x_2} \frac{\partial \phi_i(\x)}{\partial x_1}  & (\frac{\partial \phi_i(\x)}{\partial x_2})^2 & \dots  &      \frac{\partial \phi_i(\x)}{\partial x_2}\frac{\partial \phi_i(\x)}{\partial x_N} \\
      & \vdots &&\\
      \frac{\partial \phi_i(\x)}{\partial x_N} \frac{\partial \phi_i(\x)}{\partial x_1}  & \frac{\partial \phi_i(\x)}{\partial x_N} \frac{\partial \phi_i(\x)}{\partial x_2} & \dots  &      (\frac{\partial \phi_i(\x)}{\partial x_N})^2 \\
\end{bmatrix}
\]


 Smoothness here means that $\mathrm{det}(\textbf{G}(\textbf{x}))$ is higher if $\varphi(\textbf{x})$ is higher, and vise-versa. On the other hand, we make the following interesting observation. Consider the Jacobian of the embedded manifold

\[
\label{equ:jacobian}
  \textbf{J}_\Phi(\x) =
  \begin{bmatrix}
       \frac{\partial \phi_1(\x)}{\partial x_1} &  \frac{\partial \phi_2(\x)}{\partial x_1} & \dots & \frac{\partial \phi_D(\x)}{\partial x_1} \\
       & \vdots & & \\
       \frac{\partial \phi_1(\x)}{\partial x_N} &  \frac{\partial \phi_2(\x)}{\partial x_N} & \dots & \frac{\partial \phi_D(\x)}{\partial x_N} \\
\end{bmatrix}
\]

Then, the  Riemannian metric tensor $\mathcal{M}(\x) = \textbf{J}_\Phi(\x)\textbf{J}^T_\Phi(\x)$ of the manifold created by the positional embedding is equivalent to $\textbf{G}$ given in Eq.~\ref{equ:metric}. Hence, choosing $\varphi(\textbf{x}) = \|\textbf{J}_f(\x)\|_F$ and minimizing $\tau({\textbf{u}})$   encourages our embedded manifold to adhere to the constraint~\ref{equ:prac_condition}. This is a crucial observation that motivates our work. That is, constructing a graph using embedded positions as vertices, and applying the Laplacian regularization against the Jacobian norm of the encoded signal approximately fulfills  \ref{equ:prac_condition}. 
\vspace{-1em}
\section{Radial basis embedders}
\label{sec:supergaussian}
 \vspace{-0.6em}
In this section, we propose a novel trainable positional embedding scheme based on the radial basis functions that can benefit from the optimization framework developed thus far. Using the radial basis functions to encode positions were partially explored in \cite{zheng2021rethinking}. However, in the original form, their extension to higher dimensions is problematic (see Appendix) and yield inferior performance to RFF. In contrast, the proposed embedders do not suffer from such a drawback. Due to superior empirical performance, we choose a super-Gaussian as our basis function. Next, we define and discuss super-Gaussian embedders.

\begin{definition}
Given an $N$-dimensional input coordinate $\textbf{x} = [x_1, \dots, x_N]^T$, the $D$-dimensional super-Gaussian positional embedder takes the form:

\[
    \Phi(\x) = [\phi_1(\x), \phi_2(\x), \dots, \phi_D(\x)]^T,
\]

where $phi_i(\textbf{x}) = \big[ e^{-\frac{( \textbf{x} \cdot \mathbf{\alpha}- t_i)^2}{2\sigma_{\textbf{x}}^2}} \big]^b.$ Here, $t_i$ and $b$ are constants, $\mathbf{\alpha} \in \mathbb{R}^N$ is a vector with constant elements, and $\sigma_{\textbf{x}}$ is a coordinate-dependant parameter.
\end{definition}
 
 Now, we discuss 
 optimizing the super-Gaussian positional embedder using the graph Laplacian regularizer. Precisely, our aim is to obtain a coordinate dependant set of standard deviations $\{\sigma_{\textbf{x}}\}$, which minimizes  \ref{equ:glr}, and thereby approximates  \ref{equ:prac_condition}. We begin by affirming that the super-Gaussian embedders indeed form a  manifold with its local coordinate chart as the input space. Formally, we present the following proposition.

\begin{proposition}
Let the positional embedding be $\Phi(\textbf{x}) = [\phi_1(\textbf{x}) \, \phi_2(\textbf{x}) \, \dots  \phi_D(\textbf{x})]^T$ where $\phi_i(\textbf{x}) = \big[ e^{\frac{(\textbf{x} \cdot \mathbf{\alpha} - t_i)^2}{\sigma^2}} \big]^b$ for $\textbf{x} \in \RB^N$, $t_i$ and $b$ are constants, and $\mathbf{\alpha} \in \mathbb{R}^N$ is a vector with constant elements. Then, $\Phi$ is a $N-$dimensional manifold in an ambient $\RB^D$ space.
\end{proposition}

(Proof in Appendix). Then, we construct a similarity graph as discussed in Section~\ref{sec:discrete} using the embedded vectors. Afterward, one can optimize $\sigma_{\textbf{x}}$ to minimize \ref{equ:glr} using the gradient descent. However, this can lead to unexpected trivial solutions as discussed next.

\noindent Let $\mathcal{E}$ be the set of edges of our graph $\mathcal{G}$. It can be shown that,

\begin{equation}
    \tau(\textbf{u}) = \textbf{u}^T\textbf{L}\textbf{u} = \sum_{(i,j)\in \mathcal{E}}w_{i,j}( \varphi(\textbf{x}_i) - \varphi(\textbf{x}_j) )^2
\end{equation}

(see Appendix). Therefore, if $\tau(\textbf{u})$ is minimized in an unconstrained setting, the network can converge to a trivial solution by minimizing $w_{i,j}$ without preserving any graph structure. In other words, the network seeks to put the embedded coordinates as further apart as possible by decreasing $\sigma_{\textbf{x}}$ everywhere.



(Proof in Appendix). To avoid the above trivial solution, we add a second term to the objective function as,

\begin{equation}
\label{equ:norm}
    \bar{\tau}(\textbf{u}) = \textbf{u}^T\textbf{L}\textbf{u} - \lambda \|\textbf{A}\|_F,
\end{equation}

where $\textbf{A}$ is the adjacency matrix and $\lambda$ is a constant weight. Intuitively, the second term of Eq.~\ref{equ:norm} forces $\mathcal{G}$ to minimize the edge weights only at necessary points and keep other edge weights high. 


\vspace{-1em}
\subsection{Scalability}
\begin{figure}
\centering
\begin{minipage}[t]{\dimexpr 0.5\textwidth-0.4\columnsep}
\centering
    \includegraphics[width=0.8\columnwidth]{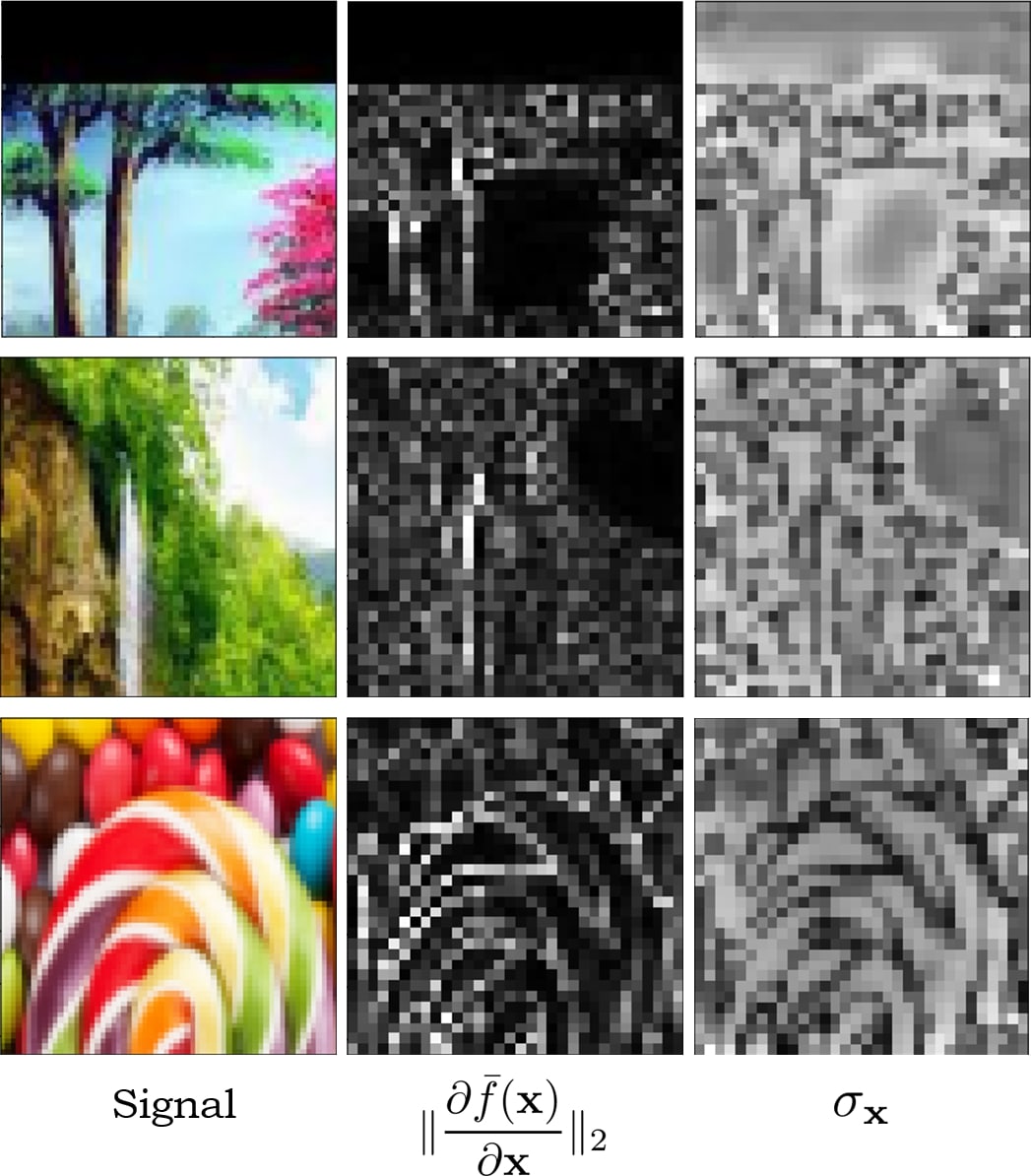}
    \caption{\small \textbf{$\| \frac{\partial \bar{f}(\x)}{\partial \x}\|_2$ vs the trained standard deviations $\sigma_{\x}$ where $\bar{f}(\x)$ is the mean of the encoded signal at $\x$.} The standard deviations are \emph{almost} a point-wise property of $\| \frac{\partial \bar{f}(\x)}{\partial \x}\|_2$. This enables us to approximate the standard deviations for unseen signals using a polynomial equation without training each time.  We use $16 \times 16$ patches for this example to clearly visualize the pixel-wise correlations.}
    \label{fig:derivatives}
\end{minipage}%
 \hfill
\begin{minipage}[t]{\dimexpr 0.5\textwidth-0.4\columnsep}
\centering
    \includegraphics[width=1.0\columnwidth]{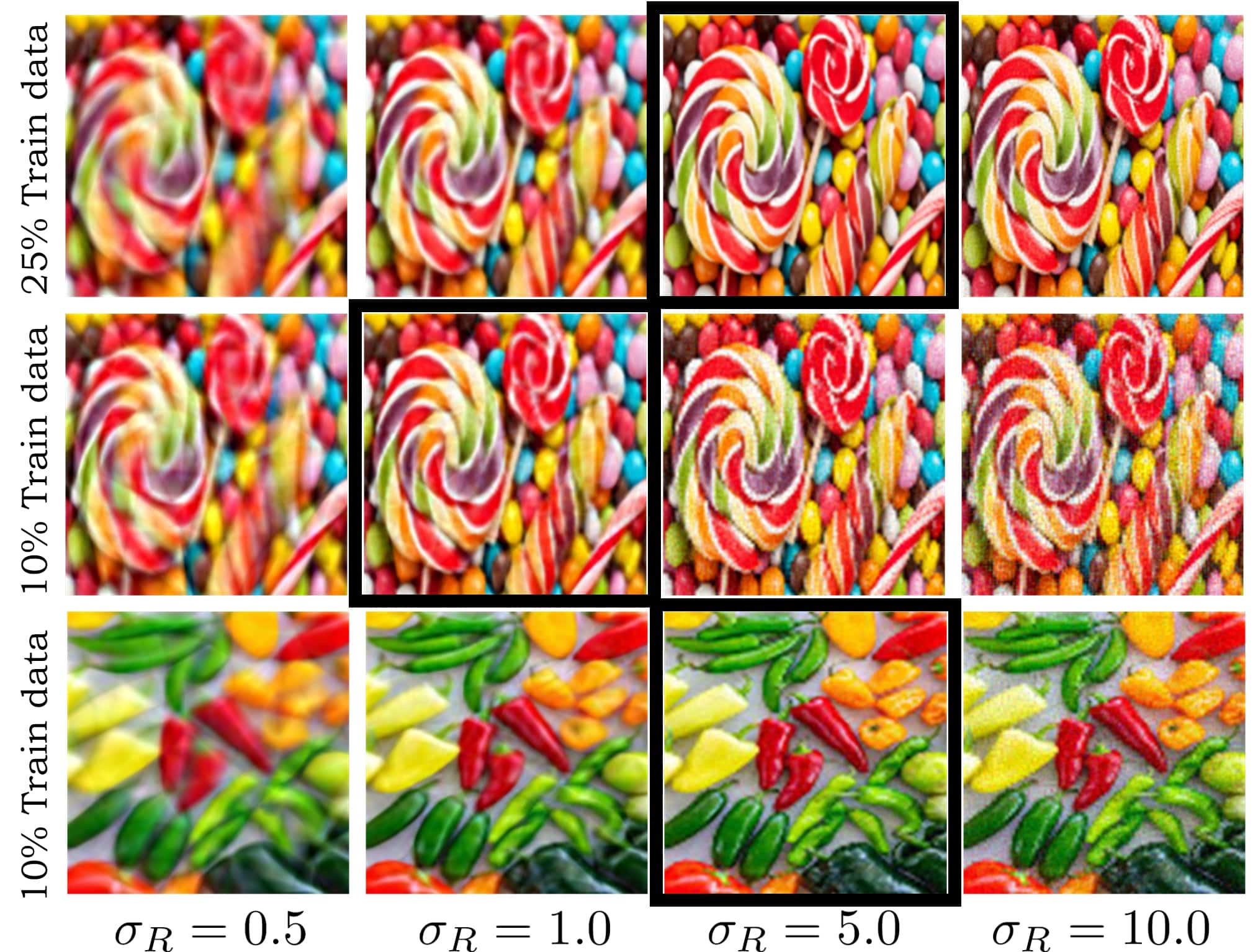}
    \caption{\small \textbf{Encoding signals with RFF (zoom-in for a better view).} The representations with the highest quality are highlighted with a black border in each row. $\sigma_R$ denotes the standard deviation of the distribution where the frequency components for RFF are chosen from. As shown, the optimal hyperparameters for RFF vary according to the characteristics of the encoded signal and the sampling procedure, making manual hyperparameter sweeps expensive and ambiguous. In contrast, our method do not suffer from such a drawback.}
    \label{fig:variance_rff}
\end{minipage}
\end{figure}

 \vspace{-0.6em}
A drawback entailed with optimizing $\sigma_{\textbf{x}}$ as described in Section~\ref{sec:methodology} is that it requires computing the adjacency matrix $\textbf{A} \in \RB^{N \times N}$, which is not scalable to large $N$. Therefore, we propose an alternative mechanism to compute $\sigma_{\textbf{x}}$ efficiently. To this end, we make a key observation that optimal $\sigma_{\textbf{x}}$  approximately depends on $\| \frac{\partial \bar{f}(\x)}{\partial \x}\|_2$, where $\bar{f}(\x)$ is the mean value of the signal output $f(\x)$. To empirically validate this, we visualize the derivative map of the signal against  $\sigma_{\textbf{x}}$. Fig.~\ref{fig:derivatives} visualizes the results. As evident, $\sigma_{\textbf{x}}$ is approximately a point-wise property of the derivative map. Leveraging the above observation, we approximate $\sigma_{\textbf{x}}$ using the following polynomial equation:

\begin{equation}
\label{equ:sigma_analytic}
    \sigma_{\x_i} = \beta_1 + \beta_2 \| \frac{\partial \bar{f}(\x_i)}{\partial \x_i}\|_2 + \beta_3 \| \frac{\partial \bar{f}(\x_i)}{\partial \x_i}\|_2^2 + \dots + \beta_L \| \frac{\partial \bar{f}(\x_i)}{\partial \x_i}\|_2^{L-1},
\end{equation}

 The coefficients $\{\beta_i\}_{i=1}^L$ of the Eq.~\ref{equ:sigma_analytic} can be found analytically, or via a linear MLP, using a small training set ($\sim 20$ samples of $\| \frac{\partial \bar{f}(\x_i)}{\partial \x_i}\|_2$ and $\sigma_{\textbf{x}_i}$ pairs). Once computed, $\{\beta_i\}_{i=1}^L$ can be used universally across similar signal classes, \textit{e.g.}, same coefficients can be used to find $\sigma_{\textbf{x}}$ for an arbitrary image. We also observe that $L=10$ is enough to provide adequate results.

\begin{figure}[!htp]
    \centering
 \includegraphics[width=0.7\columnwidth]{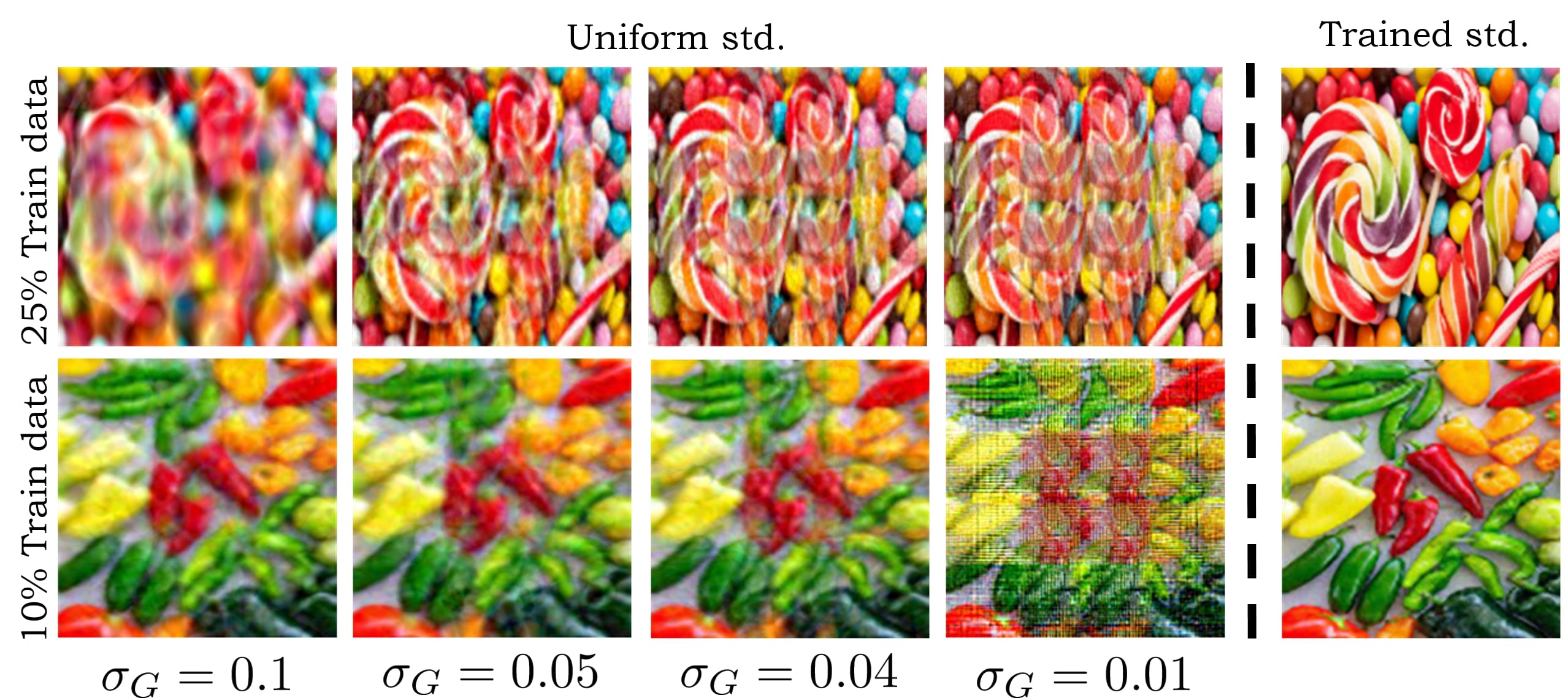}
    \vskip -0.1in
    \caption{\small \textbf{Uniform standard deviations vs trained standard deviations for encoding signals with super-Gaussian embedders (zoom-in for a better view).} Note that the different regions of the images show different reconstruction qualities with uniform standard deviations, indicating that they should be optimized per-coordinate. In contrast, the proposed training method achieves better and consistent representation quality across the image.}
    \label{fig:variance_gau}
\end{figure}

   



 \vspace{-0.6em}
\section{Experiments}
\label{sec:experiments}
 \vspace{-0.6em}
In this section, we conduct experiments to validate the efficacy of the proposed positional embedding layer and the optimization method.
 \vspace{-0.6em}
\subsection{Hyperparameters and experimental settings}
 \vspace{-0.6em}
We first construct three small datasets ($20$ samples each) of 1D, 2D and 3D signals, and then obtain the pairs $(\| \frac{\partial \bar{f}(\x_i)}{\partial \x_i}\|_2, \sigma_{\textbf{x}_i})$ by minimizing the  \ref{equ:norm}. Then, we calculate the coefficient sets $\{\beta\}_{i=1}^{L}$ of Eq.~\ref{equ:sigma_analytic}  for 1D, 2D, and 3D signals individually using least square optimization. These coefficients are then used throughout all the experiments to find $\sigma_{\textbf{x}}$ of a given signal using Eq. \ref{equ:sigma_analytic}. However, in most cases, it is required to regress to unseen coordinates. In such scenarios, we obtain the approximate first-order gradients for the sparse training points, and calculate the training $\sigma_{\textbf{x}_{train}}$. Then, $\sigma_{\textbf{x}_{test}}$ for the test coordinates are determined via linear interpolation.

The hyperparameters for the Gaussian embedder and RFF were chosen by manual hyperparameter sweeps, \textit{i.e.} for each signal, we conducted an extensive grid search for optimal hyperparameters. Note that this re-confirms the efficacy of our method. Further, all the MLPs use ReLU activations, and are trained using the Adam optimizer with a learning rate of $1 \times 10^{-4}$ and a weight decay of $1 \times 10^{-8}$.

\begin{figure}
\centering
\begin{minipage}[t]{\dimexpr 0.5\textwidth-0.4\columnsep}
\centering
\resizebox{1.\columnwidth}{!}{
\begin{tabular}{||c|c|c||}
\hline
Embedding  & Train PSNR& Test PSNR \\
\hline
\hline
No PE  & 20.42 & 20.39   \\
RFF (matched) & 34.53 & 26.03 \\
RFF (unmatched) & 32.16 & 23.24  \\
super-Gaussian (uni.)  & 34.52 & 27.19\\
super-Gaussian (e.tr.)  &  39.12 & 21.44 \\
super-Gaussian ($\beta$) & 34.59 & \textbf{31.17} \\
super-Gaussian (analytic) & 34.70 & \textbf{31.17} \\
\hline
\end{tabular}}
 \vspace{-0.7em}
  \caption{\textbf{Quantitative comparison in 1D signal encoding.} Our method outperforms  RFF. The performance of RFF drops when the parameters are not found via cross-validation per signal. (uni.), (e.tr.), ($\beta$) and (analytic) denotes uniform, end-to-end trained, proposed method with Eq~\ref{equ:sigma_analytic}, and proposed method with analytical $\sigma_\textbf{x}$ respectively.}
\label{tab:1d}
\end{minipage}%
 \hfill
\begin{minipage}[t]{\dimexpr 0.5\textwidth-0.4\columnsep}
\centering
       \begin{tabular}{||c|c|c||}
\hline
Embedding & PSNR & SSIM \\
\hline
\hline
RFF (matched) & 31.13 & 0.947\\
RFF (unmatched $\uparrow$) & 29.16 & 0.911\\
RFF (unmatched $\downarrow$) & 25.11 & 0.801\\
super-Gaussian (uni.) & 25.41& 0.891\\
super-Gaussian (p. tr.) & \textbf{32.17} & \textbf{0.981}\\
\hline
\end{tabular}
 \vspace{-0.7em}
  \caption{\textbf{Quantitative comparison embedders in 3D signal encoding on Realistic Synthetic dataset \cite{mildenhall2020nerf}}. Our embedder yields the best PSNR and SSIM in view synthesis. The performance of RFF drops When the number of unique frequency components differs from the optimal value. }
\label{tab:3D_quantitative}
\end{minipage}
\end{figure}


\begin{figure}
\centering
\begin{minipage}[b]{\dimexpr 0.5\textwidth-0.4\columnsep}
\centering
\resizebox{1.\columnwidth}{!}{
  \begin{tabular}{||c|c|c|c||}
\hline
Embedding & Image type & Train PSNR & Test PSNR \\
\hline
\hline
No PE &	Natural & 20.42 & 20.39 \\
RFF (matched) & Natural & 34.51 & 26.47 \\
RFF (unmatched) & Natural & 31.51 & 24.69 \\
super-Gaussian (uni.) & Natural & 33.22 & 26.15  \\
super-Gaussian (e. tr.) & Natural & 42.72 & 20.11 \\
super-Gaussian ($\beta$) & Natural & 36.63 & 28.17\\
Super-Gaussian (analytic) & Natural & 36.79 & \textbf{29.01}\\
\hline 
\hline
No PE & Text & 18.49 & 18.49\\
RFF (matched) & Text & 38.41 & 31.73 \\
RFF (unmatched) & Text & 36.16 & 29.54 \\
super-Gaussian (uni.) & Text & 36.93 & 30.65  \\
super-Gaussian (e. tr.) & Text & 43.88 & 23.81 \\
super-Gaussian ($\beta$) & Text & 39.76& 33.01\\
super-Gaussian (analytic) & Text & 40.11& \textbf{34.05}\\
\hline
\hline
No PE & Noise & 10.82 & 10.78 \\
RFF (matched) & Noise & 17.60 & 9.20 \\
RFF (unmatched) & Noise & 17.61 & 9.13 \\
super-Gaussian (uni.) & Noise & 11.41 & 10.55 \\
super-Gaussian (e. tr.) & Noise & 17.19 & 8.73 \\
super-Gaussian ($\beta$) & Noise & 16.31 & \textbf{12.49}\\
super-Gaussian (analytic) & Noise & 16.11 & 12.13\\
\hline
\end{tabular}}
 \vspace{-0.1em}
  \caption{\textbf{Quantitative comparison of embedders in 2D signal encoding.} Our embedder yields best PSNRs. The performance of RFF drops when the parameters are not found via cross-validation per signal. (uni.), (e.tr.), ($\beta$) and (analytic) denotes uniform, end-to-end trained, proposed method with Eq~\ref{equ:sigma_analytic}, and proposed method with analytical $\sigma_\textbf{x}$ respectively.}
\label{tab:2d}
\end{minipage}%
 \hfill
\begin{minipage}[b]{\dimexpr 0.5\textwidth-0.4\columnsep}
\centering
         \includegraphics[width=0.99\columnwidth]{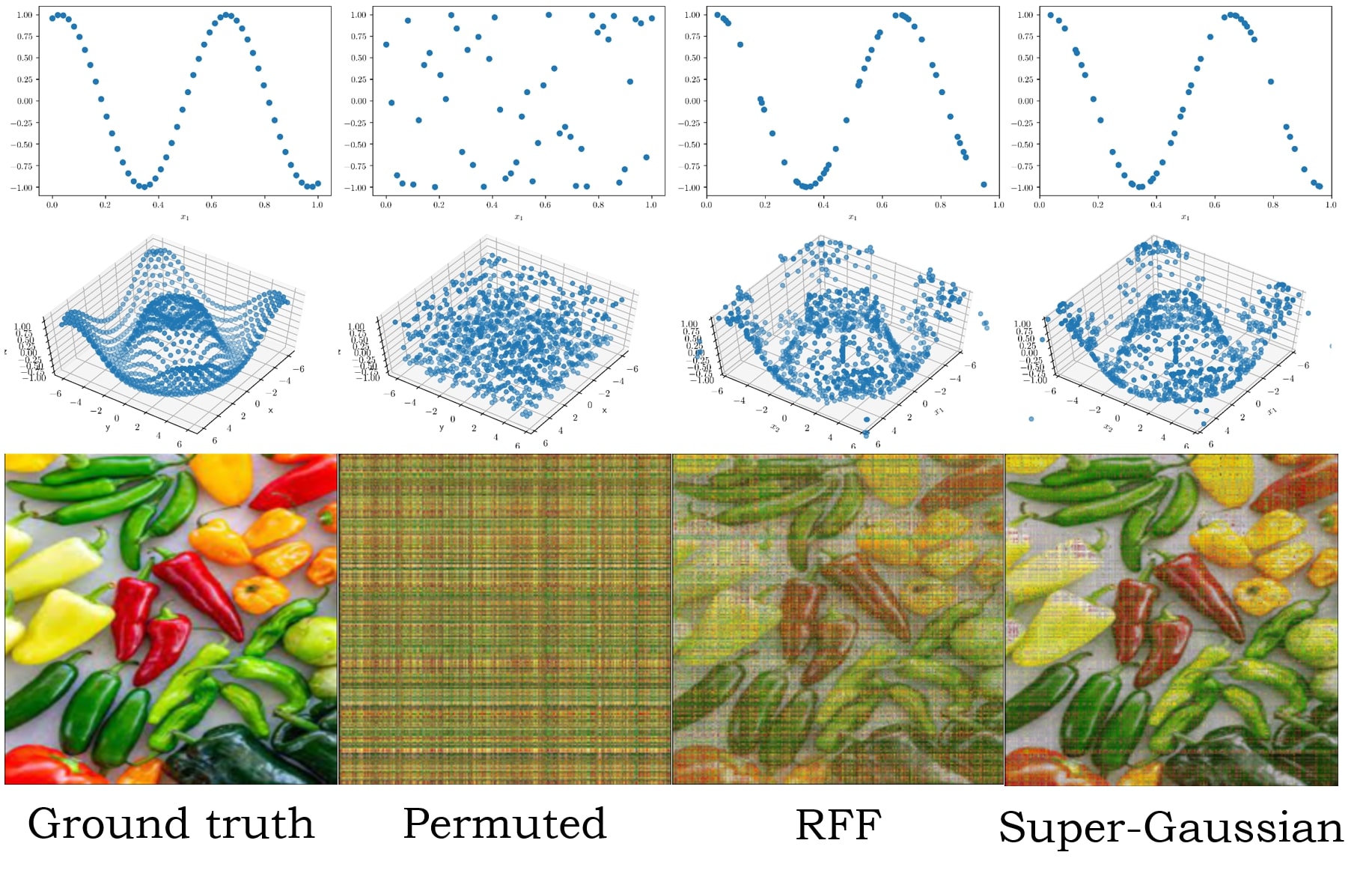}
    \caption{\small \textbf{Stable gradient flow across the embedding layer.} Each row corresponds to different signals with increasing complexity (from top to bottom). First, the MLPs are trained to overfit the signal. Then, the coordinates are permuted. Finally, the MLPs try to recover the original coordinates using gradient descent with the original image as the ground truth. As the signal complexity increases, super-Gaussian  emebdder recovers the coordinates better compared to RFF. However, the original coordinates may not be recovered exactly, as the outputs  are not unique per coordinate. }
    \label{fig:backprop}
\end{minipage}
\end{figure}

     

\begin{figure}
\centering
\begin{minipage}[b]{\dimexpr 0.5\textwidth-0.4\columnsep}
\centering
\resizebox{1.\columnwidth}{!}{
\begin{tabular}{||c|c|c||}
\hline
Embedding  & Train PSNR& Test PSNR \\
\hline
\hline
No PE  & 20.42 & 20.39   \\
RFF (matched) & 34.53 & 26.03 \\
RFF (unmatched) & 32.16 & 23.24  \\
super-Gaussian (uni.)  & 34.52 & 27.19\\
super-Gaussian (e.tr.)  &  39.12 & 21.44 \\
super-Gaussian ($\beta$) & 34.59 & \textbf{31.17} \\
super-Gaussian (analytic) & 34.70 & \textbf{31.17} \\
\hline
\end{tabular}}
 \vspace{-0.7em}
  \caption{\textbf{Quantitative comparison in 1D signal encoding.} Our method outperforms  RFF. The performance of RFF drops when the parameters are not found via cross-validation per signal. (uni.), (e.tr.), ($\beta$) and (analytic) denotes uniform, end-to-end trained, proposed method with Eq~\ref{equ:sigma_analytic}, and proposed method with analytical $\sigma_\textbf{x}$ respectively.}
\label{tab:1d}
\end{minipage}%
 \hfill
\begin{minipage}[b]{\dimexpr 0.5\textwidth-0.4\columnsep}
\centering
       \begin{tabular}{||c|c|c||}
\hline
Embedding & PSNR & SSIM \\
\hline
\hline
RFF (matched) & 31.13 & 0.947\\
RFF (unmatched $\uparrow$) & 29.16 & 0.911\\
RFF (unmatched $\downarrow$) & 25.11 & 0.801\\
super-Gaussian (uni.) & 25.41& 0.891\\
super-Gaussian (p. tr.) & \textbf{32.17} & \textbf{0.981}\\
\hline
\end{tabular}
 \vspace{-0.7em}
  \caption{\textbf{Quantitative comparison embedders in 3D signal encoding on Realistic Synthetic dataset \cite{mildenhall2020nerf}}. Our embedder yields the best PSNR and SSIM in view synthesis. The performance of RFF drops When the number of unique frequency components differs from the optimal value. }
\label{tab:3D_quantitative}
\end{minipage}
\end{figure}

\strutlongstacks{T}
\begin{table}
\centering
\scriptsize
\begin{tabular}{||c|c|c|c||}
\hline
\multirow{2}{*}{Layers}           & \multicolumn{3}{|c|}{25\% tr. data (regular)}\\
\cline{2-4}
& RFF & \Longstack{ super-Gaussian  \\ (uni.)}   & \Longstack{ super-Gaussian  \\ ($\beta$)}  \\
\hline
1 & 16.89 & 17.40 &\textbf{19.99} \\
2& 22.13 & 20.07 &\textbf{23.01}\\
3 & 23.11 & 21.13 & \textbf{23.95}\\
\hline
\hline
\multirow{2}{*}{Layers}           & \multicolumn{3}{|c|}{10\% tr. data (regular)}\\
\cline{2-4}
& RFF & \Longstack{ super-Gaussian  \\ (uni.)} & \Longstack{ super-Gaussian  \\ ($\beta$)}  \\
\hline
1 & 15.78 & 16.17 &\textbf{18.68} \\
2 & 17.76 & 17.12 &\textbf{19.51} \\
3 & 19.50  & 18.01 &\textbf{21.12} \\
\hline
\hline
\multirow{2}{*}{Layers}           & \multicolumn{3}{|c|}{25\% tr. data (random)}\\
\cline{2-4}
& RFF &  \Longstack{ super-Gaussian  \\ (uni.)}&  \Longstack{ super-Gaussian  \\ ($\beta$)} \\
\hline
1 & 16.01 & 16.80 & \textbf{19.33} \\
2 & 19.99 & 18.44 & \textbf{22.81} \\
3 & 21.83 & 19.14 & \textbf{23.41}\\
\hline
\hline
\multirow{2}{*}{Layers}           & \multicolumn{3}{|c|}{10\% tr. data (random)}\\
\cline{2-4}
& RFF & super-Gaussian (uni.) & super-Gaussian ($\beta$)\\
\hline
1 & 13.17 & 14.91 & \textbf{17.64}\\
2 & 16.41 & 15.87 & \textbf{18.32}\\
3 & 18.47 & 17.11 & \textbf{20.96} \\
\hline
\end{tabular}

\caption{\small \textbf{The expressiveness of the embedders tested against different training conditions.} The super-Gaussian embedder optimized using the proposed method is able to outperform RFF embedder with both $25\%$ and $10\%$ training data. In particular, trained super-Gaussian embedders perform better with a fewer number of layers. When used with uniformly distributed standard deviations, super-Gaussian embedders demonstrate poor results. All reported values are in mean PSNR. (uni.) and ($beta$) denotes uniform and proposed method with Eq.~\ref{equ:sigma_analytic}, respectively. we use $120 \times 120$ rescaled images from the natural category of \cite{tancik2020fourier}.}
\label{tab:expressiveness}
\end{table}

 \vspace{-0.7em}

\begin{figure*}[!htp]
    \centering
    \includegraphics[width=1.\columnwidth]{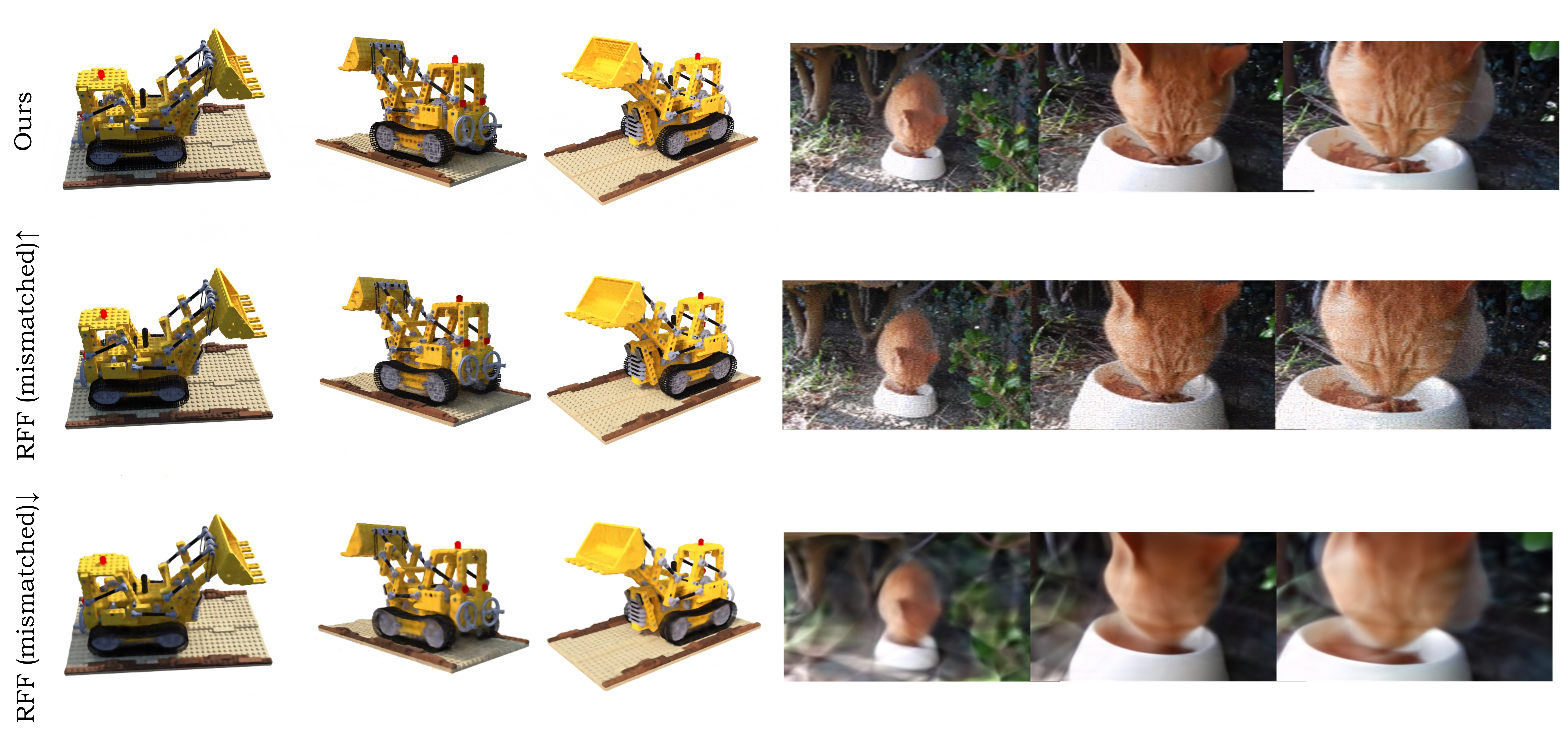}
    \vspace{-1em}
    \caption{\small \textbf{Encoding 3D signals (zoom in for a better view). } The left block shows NeRF-style \cite{mildenhall2020nerf} examples, where we used the same model as in \cite{mildenhall2020nerf} with our embedding. The right block illustrates three frames of an encoded video. When RFF deviates from the optimal parameter setting, the results are considerably inferior to the super-Gaussian embedder. Even with the optimal parameter setting, the super-Gaussian embedder achieves better quantitative results compared to RFF (Table \ref{tab:3D_quantitative}).}
    \label{fig:qualitative_3d}
\end{figure*}

 \vspace{-1.em}
\subsection{Encoding signals}
 \vspace{-0.6em}
In this section, we compare the performance of  embedders in enabling coordinate-MLPs to encode signals across various settings.

\noindent{\textbf{1D signals: }}First, we pick $25$ random rows from the natural 2D images released by \cite{tancik2020fourier} to create a small dataset of 1D signals, each with $512$ pixels. Then, we encode the coordinates using different embedding layers and feed each of them to a 4-layer MLP to encode the signal. For each signal, we sample $256$ points with an interval of one as the training set, and the rest of the points as the testing set. Experiments were repeated $10$ times to obtain the average performance for each embedder. Table~\ref{tab:1d} depicts the results. As shown, our embedder reports the best train, and test PSNRs over RFF when trained using the proposed method. In RFF (matched), we conducted a grid search to obtain the best parameters for each signal. In RFF (unmatched), we find the best parameters for the first signal, and keep it constant across all the signals. As illustrated, RFF (unmatched) yields inferior performance to the RFF (matched), which confirms that the optimal parameters for RFF indeed depend on the encoded signal. 

\noindent{\textbf{2D signals: }} For experimenting on 2D images, we use the dataset provided by \cite{tancik2020fourier}. It consists of three image types: Natural, Text, and Noise. As depicted in Table \ref{tab:2d}, our embedder showcases similar advantages to the previous $1D$ signal encoding experiment.

\begin{figure*}[!htp]
    \centering
    \includegraphics[width=1.\columnwidth]{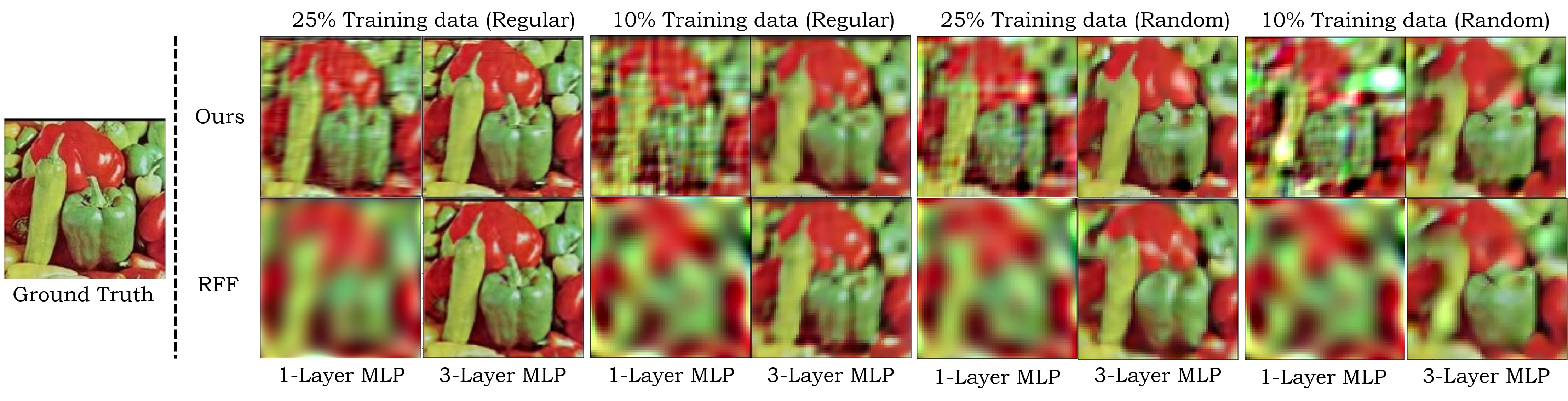}
     \vspace{-1.em}
    \caption{\small\textbf{ A qualitative comparison of the super-Gaussian embedder against RFF in 2D image representations.} The super-Gaussian embedder achieves significantly better results with a linear MLP compared to RFF under various sampling conditions. However, as deeper MLPs with non-linearity are used, RFF representations rapidly increase in quality, although our method consistently produces better results ( Table \ref{tab:expressiveness}). We used $128 \times 128$ images for this example for clear visual contrast.}
    \label{fig:qualitative_2d}
\end{figure*} 

\noindent{\textbf{3D signals: }} For evaluating the embedders on 3D signals, we utilize NeRF-style 3D scenes. The quantitative results are shown in Table \ref{tab:3D_quantitative}. RFF (matched) uses log linear sampling of frequencies with the maximum frequency $2^{10}$. RFF (unmatched $\uparrow$) and  RFF (unmatched $\downarrow$) use $2^{15}$ and $2^{8}$ as the maximum frequency component, respectively.  When the chosen frequency components deviate from the ideal values, RFF demonstrates a drop in performance. This behavior confirms the importance of a principled method for finding the optimal parameters for a positional embedding.

 Fig.~\ref{fig:variance_rff} depicts a qualitative example that further validates the high cost of using cross-validation to obtain optimal parameters for RFF.  We test the encoding performance of coordinate-MLPs, when equipped with RFF positional embeddings with varying hyperparameters. As Fig~\ref{fig:variance_rff} illustrates, the ideal parameters depend on the signal and the sampling scheme, making the cross-validation process difficult and expensive. Therefore, extensive grid searches are required to gain optimal performance. In contrast, our embedder does not suffer from such a drawback. Fig.~\ref{fig:variance_gau} shows a qualitative comparison between uniformly distributed and trained $\sigma_{\x}$.

Across all the experiments, we compare the performance variation when using super-Gaussian embedders with uniform, end-to-end trained, and graph-Laplacian trained standard deviations. As expected, the end-to-end trained embedder yields high training PSNR, but low test PSNR. Uniform standard deviations provide a better trade-off between the two, but show inferior performance to the proposed training scheme. Across all the experiments, the super-Gaussian embedders trained with the proposed method outperform the other methods consistently.

\subsection{Expressiveness of the representation}

 Recall that a bulk of our derivations stemmed from enforcing the smoothness of the embedding layer with respect to the output. Therefore, it is reasonable to hypothesize that our embedder should demonstrate superior performance with shallow networks. To validate this, we compare the performance of the embedders while varying the depth of the MLP. As expected, we observed consistently better performance with our embedder across different depth settings (Table \ref{tab:expressiveness}). Further, we evaluate the robustness of the embedders under different sampling conditions. As illustrated in Table \ref{tab:expressiveness}, our method outperforms RFF under various sampling schemes. Fig.~\ref{fig:qualitative_2d} illustrates qualitative examples for this experiment.

\subsection{Stable gradients}

Being able to backpropagate through a module is crucial in deep learning. Stable backpropagation through a module implies that it can be integrated into a deep network as an intermediate layer. As evident from Fig.~\ref{fig:backprop}, the proposed embedder is able to produce more stable gradients compared to RFF, especially as the complexity of the signal increases.
 \vspace{-1em}

\section{Conclusion}
 \vspace{-0.6em}
The primary objective of this paper is to develop a  framework that can be used to optimize positional embeddings. We further propose a novel positional embedding scheme based on the super-Gaussian functions, which can significantly benefit from the proposed optimization strategy. We validate the efficacy of our embedder over the popular RFF embedder across various tasks, and show that the proposed super-Gaussian embedders yield better fidelity and stability in different training conditions. Finally, we demonstrate that compared to RFF, the super-Gaussian embedder is able to produce smooth gradients during backpropagation, which allows  using the embedding layers as intermediate modules in deep networks.



\bibliographystyle{splncs04}
\bibliography{main}

\newpage
\appendix
\onecolumn

\section*{\Large Appendix}
\section{Proof for Proposition 1}

\noindent \textbf{Proposition 1:} \textit{Let the positional embedding be $\Phi(\textbf{x}) = [\phi_1(\textbf{x}) \, \phi_2(\textbf{x}) \, \dots  \phi_D(\textbf{x})]^T$ where $\phi_i(\textbf{x}) = \big[ e^{\frac{(\textbf{x} \cdot \mathbf{\alpha} - t_i)^2}{\sigma^2}} \big]^b$ for $\textbf{x} \in \RB^N$, $t_i$ and $b$ are constants, and $\mathbf{\alpha} \in \mathbb{R}^N$ is a vector with constant elements. Then, $\Phi$ is a $N-$dimensional manifold in an ambient $\RB^D$ space.}

The proof is straightforward. Observe that ($\frac{\partial \Phi(\x)}{\partial \x}$) exists for all $\x$. Also, $\Phi(\x_1) = \Phi(\x_2) \iff \x_1 = \x_2$. Therefore, $\Phi$ is a continuous bijection. Further, the space $\Phi$ is a Hausdorff space and the domain of $\Psi$ is compact. Recall the following theorem.

\noindent \textbf{Theorem: } Continuous bijection from a compact space to a Hausdorff space is a homeomorphism.

Therefore, $\Phi$ is a $N$-D manifold and its local coordinate chart is a compact subspace of $\mathbb{R}^N$.

\section{Signal encoding}
\begin{figure*}
    \centering
    \includegraphics[width = 1.\columnwidth]{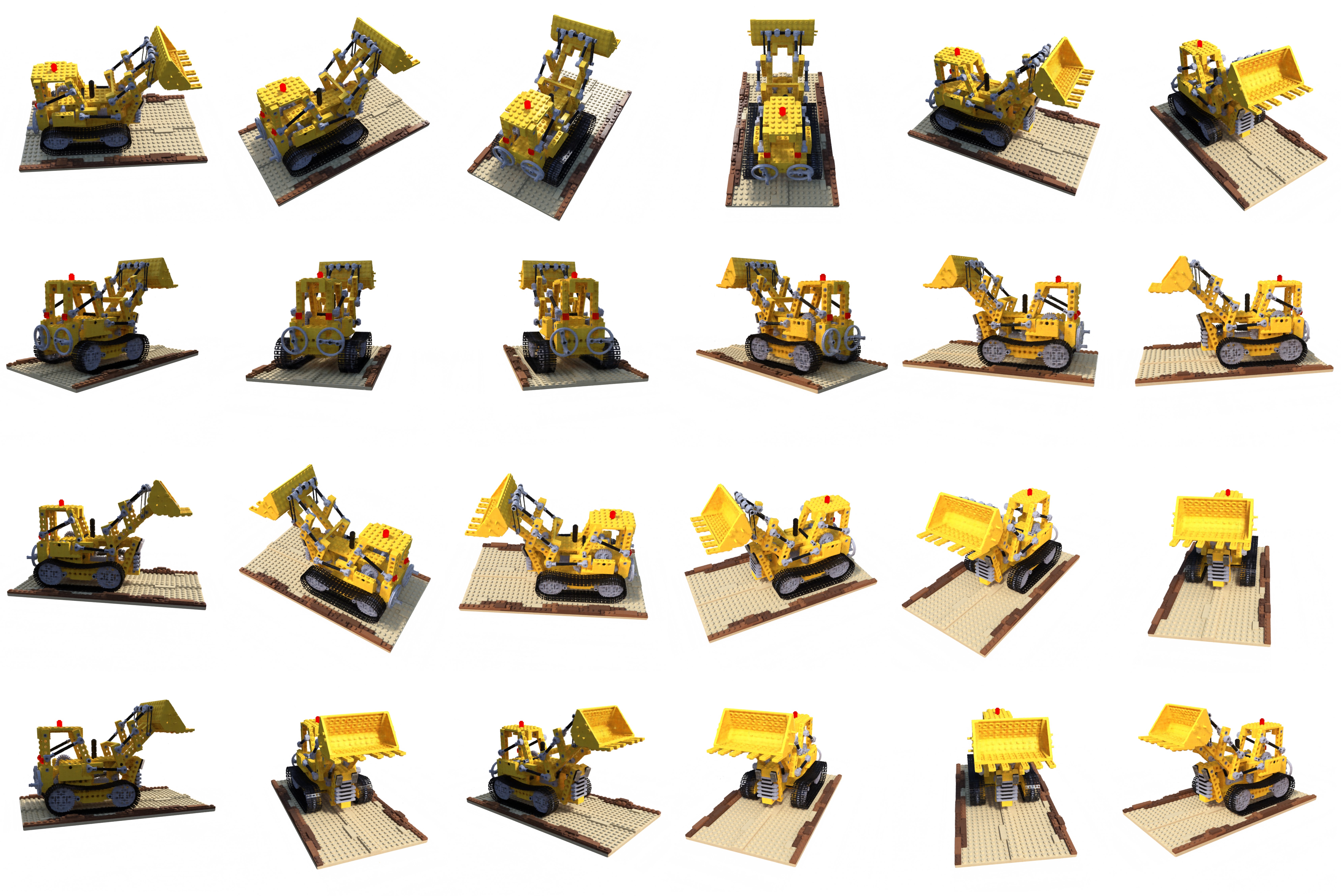}
    \caption{Qualitative examples of novel view synthesis.}
    \label{fig:lego}
\end{figure*}

\begin{figure*}
    \centering
    \includegraphics[width = 1.\columnwidth]{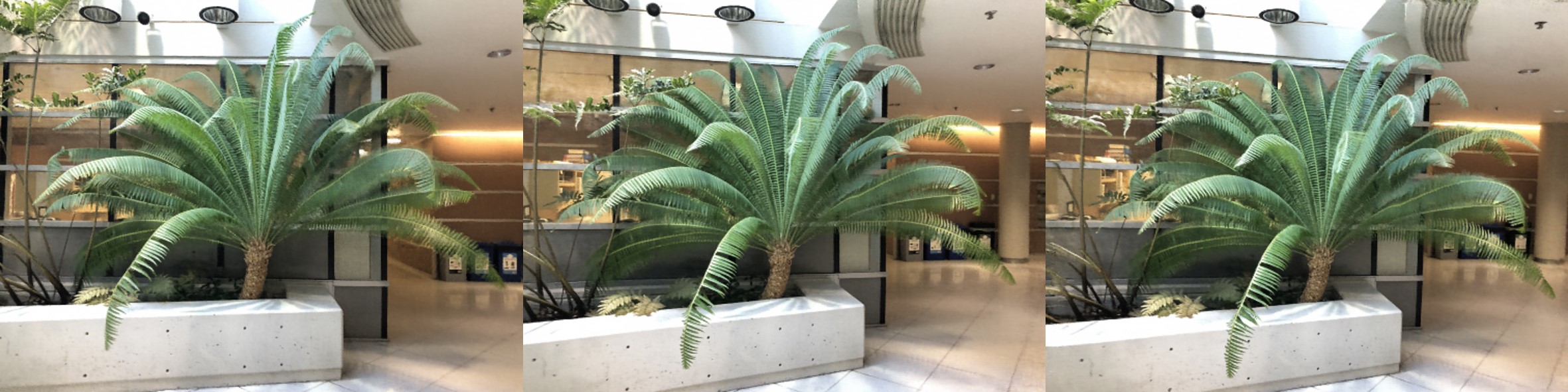}
    \caption{Qualitative examples of novel view synthesis.}
    \label{fig:lilies}
\end{figure*}

\begin{figure*}
    \centering
    \includegraphics[width = 1.\columnwidth]{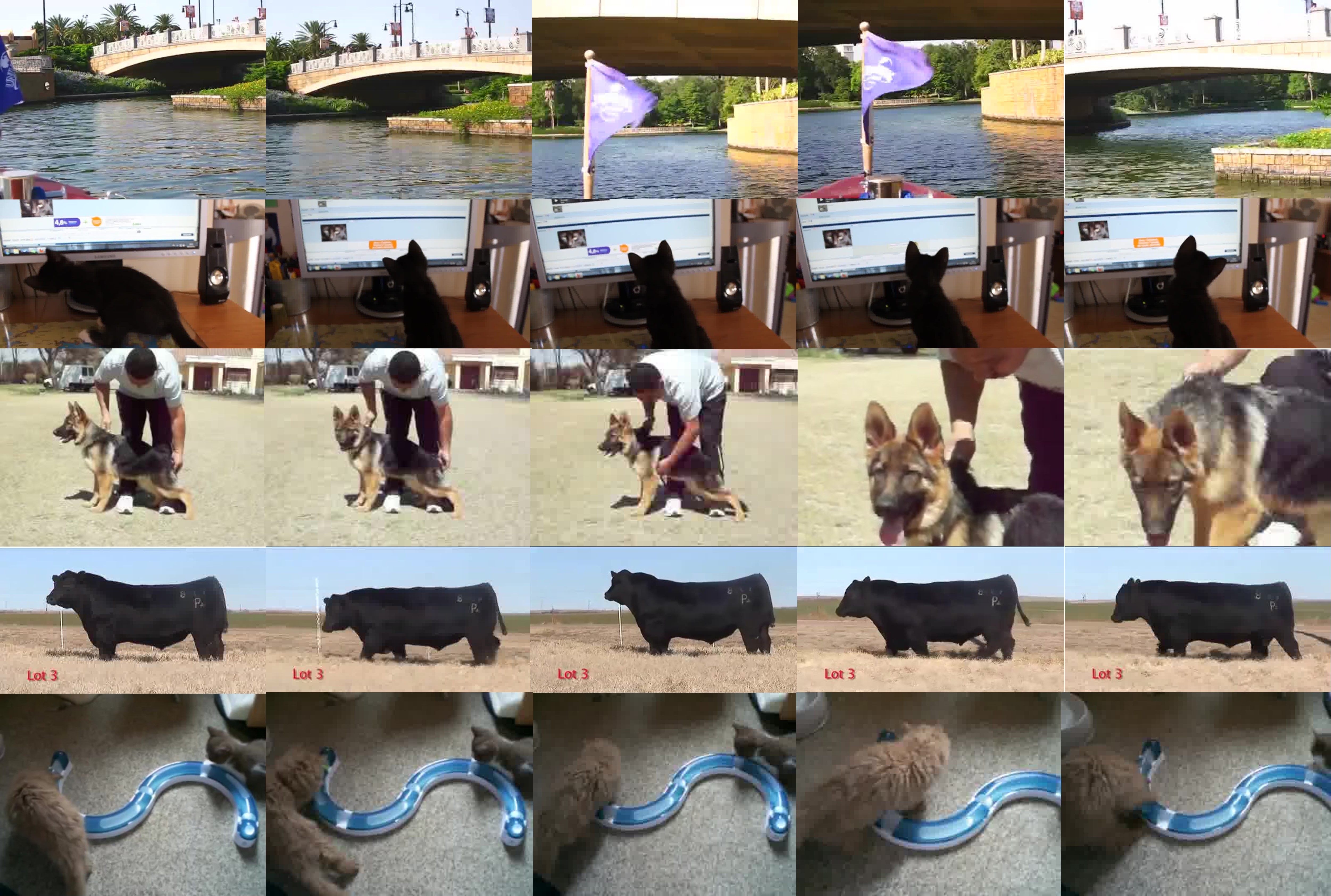}
    \caption{Random frames from the encoded videos in YouTube bounding boxes dataset.}
    \label{fig:pe}
\end{figure*}

\begin{figure*}
    \centering
    \includegraphics[width = 1.\columnwidth]{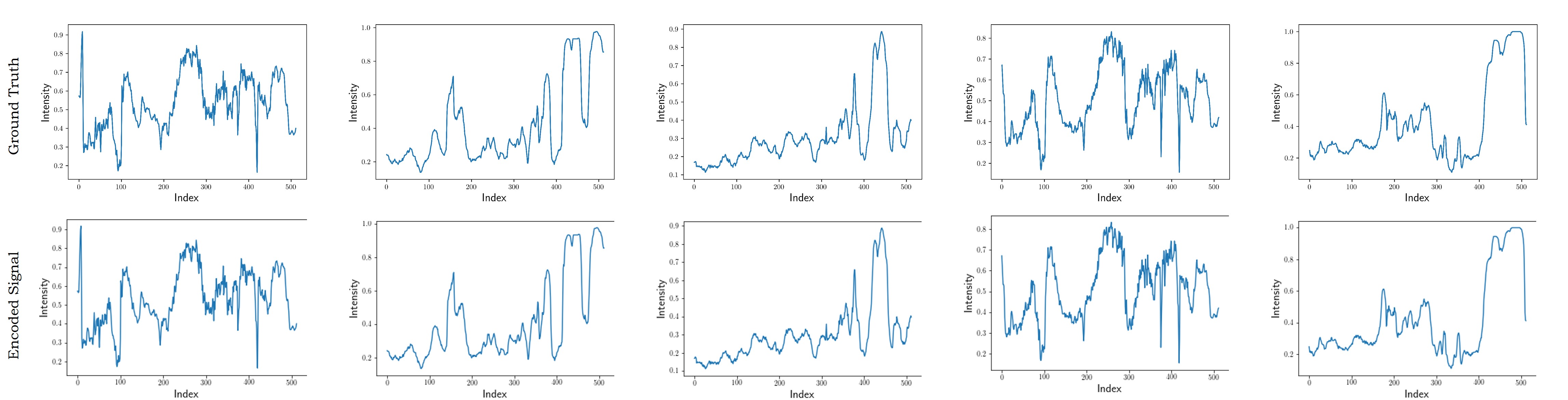}
    \caption{1D signal encoding using the super-Gaussian embedder. The examples shown are pixel intensities from random rows of images by \cite{tancik2020fourier}}
    \label{fig:1d}
\end{figure*}

\end{document}